\documentclass[twoside,11pt]{article}
\usepackage{jmlr2e}
\usepackage{booktabs}
\usepackage{amsmath}
\usepackage{multirow}
\usepackage{color, colortbl}
\definecolor{Gray1}{gray}{0.7}
\definecolor{Gray2}{gray}{0.9}

\usepackage{lastpage}
\jmlrheading{}{2024}{1-\pageref{LastPage}}{1/21; Revised 5/22}{9/22}{21-0000}{Sonit Singh}

\ShortHeadings{Clinical Context-aware Radiology Report Generation}{Sonit Singh}
\firstpageno{1}

\begin{document}

\title{Designing a Robust Radiology Report Generation System}

\author{\name Sonit Singh \email sonit.singh@unsw.edu.au \\
       \addr School of Computer Science and Engineering\\
       Faculty of Engineering, University of New South Wales\\
       Kensington, NSW 2052, Australia
       \AND
       \addr School of Computing\\
       Faculty of Science and Engineering, Macquarie University\\
       Macquarie Park, NSW 2109, Australia}

\editor{My editor}

\maketitle

\begin{abstract}
Recent advances in deep learning have enabled researchers to explore tasks at the intersection of computer vision and natural language processing, such as image captioning, visual question answering, visual dialogue, and visual language navigation. Taking inspiration from image captioning, the task of radiology report generation aims at automatically generating radiology reports by having a comprehensive understanding of medical images. However, automatically generating radiology reports from medical images is a challenging task due to the complexity, diversity, and nature of medical images. In this paper, we outline the design of a robust radiology report generation system by integrating different modules and highlighting best practices drawing upon lessons from our past work and also from relevant studies in the literature. We also discuss the impact of integrating different components to form a single integrated system. We believe that these best practices, when implemented, could improve automatic radiology report generation, augment radiologists in decision making, and expedite diagnostic workflow, in turn improve healthcare and save human lives.
\end{abstract}


\section{Introduction}

With the advances in deep neural network architectures, availability of large-scale annotated datasets, and rise of graphics processing units (GPUs), the field of deep learning~\citep{Goodfellow-et-al-2016} has shown remarkable performance in areas such as computer vision, natural language processing, and speech processing. Although research has been advancing in standalone fields for many decades, there has been rise in interest in multimodal learning~\citep{Liang:2024:Foundations_MML,Singh:2018:PLR}, making models capable of processing information from different modalities, including images, videos, speech, and text. Taking inspiration from image captioning~\citep{Vinyals:2015:Show_and_Tell}, the task of generating a description of an image, radiology report generation is a multimodal task that aims to transform radiology images into reports.

\begin{figure}
\small
{%
  \begin{minipage}{0.45\textwidth}
  \centering
    \includegraphics[width=7cm, height=5cm]{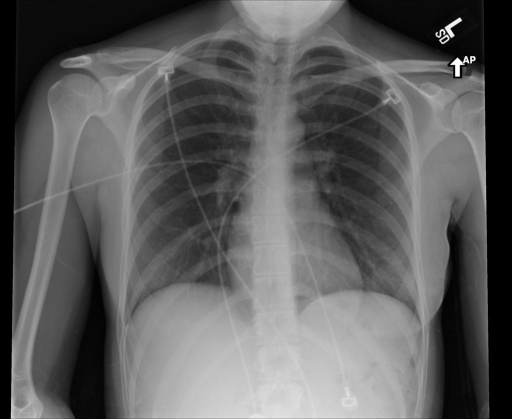}
  \end{minipage}}
\hspace{\fill} 
{%
  \begin{minipage}[c]{0.51\textwidth}
    \begin{small}
\vspace{0.5cm}
\textbf{Comparison}: \\
None. \\
\textbf{Indication}:\\
Chest pain, feels out of it. \\
\textbf{Findings}: \\
The Cardiomediastinal silhouette and pulmonary vasculature are wining normal limits in size. The lungs are clear of focal airspace disease, pneumothorax, or pleural effusion. There are no acute bony findings. \\ 
\textbf{Impression}: \\
No acute cardiopulmonary findings. \\
\end{small}
  \end{minipage}}
  \caption{An example of a chest X-ray and its corresponding radiology report in the IU-CXR dataset~\citep{Demner-Fushman:2016:IUCXR}.}
  \label{fig1:sampleReport}
\end{figure}

In radiology practice, radiologists routinely examine medical images and conclude their findings in the form of radiology reports. The radiology report indicates what abnormalities are present as well as highlights what is normal in the medical image. In Figure~\ref{fig1:sampleReport}, we show a sample chest X-ray and its corresponding radiology report from the Indiana University Chest X-ray Collection (IU-CXR)~\citep{Demner-Fushman:2016:IUCXR}. The radiology report has a structure with sections, namely, \emph{Comparison}, \emph{Indication}, \emph{Findings}, and \emph{Impression}. The task of examining medical images and writing radiology reports is \emph{tedious}, \emph{time-consuming}, \emph{subjective}, and \emph{error-prone}. With increasing population and improving imaging techniques, the demand for medical scans has been on rise. Automated radiology report generation are assistive systems that take as input the medical images and generates a textual radiology report that describes the radiological findings and interpretation. The radiology report generation system offer the potential to improve radiology practice by automating the repetitive process of radiology report drafting, identifying possible medical conditions, and reducing diagnostic errors. These systems when implemented in real clinical settings have the potential to expedite clinical workflow by triaging patients depending upon the level of urgency, assisting radiologists by providing ``second opinion", and highlighting relevant regions of interest in the images, potentially saving human lives.

\section{Related Work}

Research in radiology report generation falls under four main categories, namely \emph{template-based}~\citep{Kisilev:MRG:2015}, \emph{retrieval-based}, \emph{generation-based}, and \emph{hybrid} methods. In template-based approach, first different objects, attributes, and actions are detected from the visual content, and then blank slots are filled. \cite{Kisilev:MRG:2015} was the first to apply template based approach for automated radiology report generation where feature based approach is applied to predict categorical BI-RADS descriptors for breast lesions. Three main descriptors, namely, \emph{shape}, \emph{margin}, and \emph{density} are used to train a classifier. Further categories for each feature are: \textbf{shape}: \{\emph{oval}, \emph{round}, \emph{irregular}\}, \textbf{margin}: \{\emph{circumscribed},  \emph{indistinct}, \emph{spiculated}, \emph{microlobulated}, \emph{obscured}\}, and \textbf{density}: \{\emph{non\-homogeneous},  \emph{homogeneous}\}. Although, the trained model can identify these features given a input image and fill them into a fixed template, the approach is limited to keywords and do not support coherent and free-form radiology report. The template-based methods are simplistic in nature and can generate grammatically correct sentences due to templates, however, these methods can't generate novel sentences, nor can generate variable length sentences. Apart from this, template-based methods have rigid sentence structure. 

In retrieval-based approach, sentences are retrieved from a set of existing pool of sentences from the training data. Given an input image, first similarity between images is calculated and the caption (or sentences) of the most similar image are selected as the output for the given image. In \citep{kougia:2021:RTEX}, cosine similarity is computed between the extracted visual embeddings of images and then corresponding sentences are chosen from the pool. In \citep{ni-2020-learning}, first visual and semantic features are aligned and visual-semantic similarities are computed via an attention-weighted sum of squared l2-normalised Euclidean distance. In \citep{Li:2018:hybrid-retrieval_generation_mic}, reinforcement learning is used to train an agent to retrieve sentences from the training pool. \cite{Endo_2021_retrieval-based} used pre-trained contrastive language-image model, namely CLIP model, for retrieval-based radiology report generation. In \citep{jeong:2023:multimodalimagetextmatchingimproves}, a novel algorithm, namely, Contrastive X-ray REport Match (X-REM), was proposed that used an image-text matching score to measure the similarity of chest X-ray and radiology report for report retrieval. Although retrieval-based methods generate syntactically correct sentences, they can't generate image specific and semantically correct sentences. Apart from this, retrieval-based methods has high space search complexity due to selecting sentences from the large database or (pool of sentences) of training data.

With the rise of deep neural networks, generation methods became popular. These methods first analyse the visual content of the medical image and generate report conditioning on the visual content. Generation methods are in general capable of generating novel sentences that are semantically and syntactically more accurate than the template-based and retrieval-based methods. The most common approach for generate methods is \emph{encoder-decoder framework}, where encoder is typically a convolutional neural network (CNN) which extracts image features and the decoder is a text decoder, typically a recurrent neural network (RNN) such as Long Short-Term Memory Network (LSTM)~\citep{LSTM}, Gated Recurrent Unit (GRU)~\citep{GRU}, or Transformer~\citep{Vaswani:2017:transformer}, which learns the mapping from the image representation to text representation and consequently generates radiology report. \cite{Shin:2016:Learning_to_read_chest_X-rays} used a cascaded CNN-RNN captioning model to generate description about the detected disease. Their model could generate individual words, however the generated words are not coherent and can be difficult to comprehend. This is due to the poor information extraction and language modelling capabilities. Later \citep{Zhang:2017:MDNet} proposed a CNN-RNN model enhanced by an auxiliary attention sharpening (AAS) module to automatically generate medical imaging report. They also demonstrated the corresponding attention areas of image descriptions. Their proposed model can generate more natural sentences but the length of each sentence is limited to $59$ words. In addition, the content of a generated report was limited to five topics. \cite{Jing:2018:Automatic_Generation_of_Medical_Imaging_Reports} proposed a hierarchical co-attention based model for generating medical imaging report. Their proposed model have capability to attend image features and predict semantic tags while exploring joint effects of visual and semantic information. Also, their model can generate long sentence descriptions in reports by incorporating a hierarchical Long-Short Term Memory (LSTM) network. However, they concatenated findings and impression section into a single text for generation task. Also, the experimental results show that their model was prone to generating false positives because of the interference of irrelevant tags. In \citep{Singh:2019:from_chest_X-rays}, a CNN-LSTM model was proposed and different sort of transfer learning approaches were applied. More recently, there has been interest in leveraging large language models (LLMs) and generative artificial intelligence (GenAI). These studies~\citep{kapadnis-etal-2024-serpent, Zhang:2024, Tang:2024:generating_colloquial_reports, Zhangu:2023:R2GenGPT} have applied transformer~\citep{Vaswani:2017:transformer}, vision transformer~\citep{Dosovitskiy:2021:VIT}, LLMs such as OpenAI GPT-4~\citep{openai2024gpt4technicalreport} and Meta's Llama3~\citep{dubey2024llama3herdmodels}.

More recently, hybrid approaches has been applied, integrating the benefits of template-based, retrieval-based, and generation-based methods~\citep{Li:2018:hybrid-retrieval_generation_mic}, and use of multi-modal models~\citep{zhao2024large, srivastav-etal-2024-maira}, and knowledge graphs~\citep{Li:2019_knowledge_driven, Ranjit:2023:RAG, Li:2024:KARGEN, Parres:2024:improving_radiology}.  \cite{Li:2018:hybrid-retrieval_generation_mic} built a hierarchical reinforced agent, which introduced reinforcement learning and template based language generation method for medical image report generation. Their agent effectively utilise the benefit of retrieval and generation based approaches. However, the heavy involvement of pre-processing in extracting templates makes the method heuristic and difficult to generalise to other datasets and applications. LLMs when integrated with knowledge graphs (or knowledge bases) demonstrate significant capabilities in generating coherent, and contextually relevant responses. In \citep{Li:2024:KARGEN}, a novel Knowledge-enhanced Automated radiology Report GENeration (KARGEN) was proposed, which integrates a disease-specific knowledge base to incorporate medical knowledge. In \citep{Ranjit:2023:RAG}, retrieval augmented generation (RAG) is combined with LLM to generate radiology report generation. \cite{tang:2024:generating_clloquial} used LLM to generate colloquial translation of radiology reports by providing specialised prompts. \cite{Liu_Tian_Chen_Song_Zhang_2024} proposed a bootstrapping LLM approach having an in-domain instance induction and a coarse-to-fine decoding process to avoid mismatch between general domain text and medical text data. With the need of interpretable, explainable, and trustworthy medical systems, recently researchers have been focusing on more interactive, explainable, and trustworthy radiology report generation~\citep{Tanida:2023:Interactive_and_Explainable, Li:2021:FFA-IR, ahmed:2022:explainable_ai, Wang:2024:trust}. 
 
\section{Lessons learnt for building a robust radiology report generation}
Building a robust radiology report generation system need critical protocols which must be validated in the system. These protocols include: (1) identifying the correct use of medical concepts present; (2) incorporating medical domain knowledge into the system; (3) identifying concurrent diseases; (4) generating a coherent and long text which describes medical conditions; (5) separating different sections of radiology report such as \emph{findings} and \emph{impression} sections; and (5) evaluating the system in terms of completeness and clinical accuracy. In the following sections, we outline best practice from various studies~\citep{Singh:2018:modality_classification,Singh:2019:from_chest_X-rays,Singh:2021:show_tell_summarise} and relevant literature on radiology report generation to build robust radiology report generation systems having capability to adhere the above listed points. We first highlight the salient findings and finally provide a complete integrated system. A radiology report generation system could be improved by:

\begin{enumerate}
    \item \emph{Pre processing applying concept detection to the CNN output}
    
    A radiologist interprets medical images and describes the findings in the form of a textual radiology report outlining various medical concepts, their location, and their severity. Adding a concept detector for medical images is analogous to object detection for general images. Applying a concept detector first to find all relevant concepts present in a chest X-ray and passing the embeddings of these concepts to the text generation module is beneficial as is evident from studies~\citep{Singh:2018:modality_classification,Singh:2019:from_chest_X-rays,jing:2018:on_the_automatic,Alfarghaly:2021:Automated_radiology_report,Yuan:2019:Automatic_radiology_report}. 
    
    \item \emph{Pre training the language model on a large scale radiology corpus rather than a general corpus}
    
    \emph{Transfer learning}~\citep{Razavian:2014:Transfer_Learning} transfers knowledge learned from large-scale datasets to improve learning in related target tasks. Transfer learning plays a vital role in the medical domain where collecting large-scale annotated data is difficult due to interoperability, privacy, legal, and economical issues. In \citep{Singh:2019:from_chest_X-rays}, authors investigated the effect of general word embeddings, namely \emph{Glove}~\citep{Pennington:2014:glove}, which are trained on a generic text corpus, namely Common Crawl. They also applied \emph{Radglove} (radiology glove)~\citep{Zhang:2018:RadGlove} embeddings, which are trained on $4.5$ million radiology reports at Stanford University. The experimental results shows that applying medical imaging domain knowledge by initialising the decoder with \emph{radglove} embeddings provide better results compared to general text \emph{Glove} embeddings. This highlights the importance of incorporating medical domain knowledge into the developed system. 
    
    \item \emph{Pre training the CNN on large scale radiology imaging data rather than on general image data}
    
    Most of the proposed radiology report generation systems are based on \emph{encoder-decoder} framework, where the encoder is a CNN and decoder is a LSTM or state-of-the-art language model such as the transformer~\citep{Vaswani:2017:transformer}. \cite{Singh:2018:modality_classification} enabled knowledge transfer for extracting image embeddings using CNNs with pre-trained weights from large-scale image classification dataset such as the ImageNet for the modality classification and concept detection tasks. In~\citep{Singh:2019:from_chest_X-rays}, authors again initialised the CNN using transfer learning for detecting thoracic diseases from chest X-rays. In~\citep{Singh:2021:show_tell_summarise}, authors initialised the CNN as an encoder with pre-trained embeddings for radiology report generation. The experimental results based on proposed methods indicate that pre-trained CNN models on large-scale imaging datasets are beneficial compared to the random initialisation.
    
    \item \emph{Using separate report generators for normal versus abnormal concepts}
    
    In clinical practice, radiologists generally follow a fixed template for writing radiology reports. Most of the radiology reports consists of sections, such as, \emph{Comparison}, \emph{Indication}, \emph{Findings}, and \emph{Impression}. Out of these sections, findings and impression section are the most important sections. The \emph{findings} section indicates whether the appearance of each area in an image is normal or abnormal, and if abnormal, provides a description and suggests possible causes. The \emph{impression} section summarises the findings, patients clinical history, indication of study, and is considered to be a conclusive text for decision making. In~\citep{Singh:2021:show_tell_summarise}, authors analysed the text of reports and classifying them as either \emph{normal} or \emph{abnormal}. Based on the analysis, they find that abnormal reports are longer than normal reports as radiologists provide a detailed description outlining all abnormalities and their salient regions. On the other hand, if report is normal, radiologists generally don't write detailed description and simply summarise report as normal. Given the different nature of text for normal and abnormal radiology reports, in~\citep{Singh:2021:show_tell_summarise}, authors proposed \emph{encoder-decoder} framework separating the generation of \emph{normal} and \emph{abnormal} radiology reports. The experimental results validated the hypothesis and proved that separating \emph{normal} and \emph{abnormal} radiology report generation was of benefit to the overall radiology report generation system.  
    
    \item \emph{Generating findings from the concept generator output and then summarising to form the impression}
    
    Most of the studies of radiology report generation focused on generating the concatenation of the findings and the impression sections. Inspired by the fact that radiologists first interpret medical images and write the \emph{findings} section to highlight abnormalities present in the image. Based on the findings section along with background context in terms of history and demographics of a patient, radiologists briefly summarise findings into an \emph{impression} section which concludes the report. In~\citep{Singh:2021:show_tell_summarise}, authors relaxed the limitations of concatenating the findings and impression section as the target report to be generated. They proposed a \emph{show, tell, and summarise} model which firstly classifies whether an input chest X-ray is \emph{normal} or \emph{abnormal}, secondly the findings generation module generates \emph{findings}, and lastly the summarisation module summarises the generated findings into an \emph{impression} section. Experimental results proved the hypothesis that first generating the findings section from an input chest X-ray and then obtaining an impression section by summarising findings is beneficial to the building of a robust radiology report generation system.  
    
    \item \emph{Using state-of-the-art language models such as transformer rather than the conventional LTSM} 
    
    Recently, state-of-the-art language models, such as the transformer~\citep{Vaswani:2017:transformer} has shown remarkable progress on various language understanding and generation tasks. Taking inspiration from this, ~\cite{singh:2024:clinicalcontextawareradiologyreport} applied a transformer model based radiology report generation, which replaces the recurrent architecture with a self-attention mechanism. Experimental results show that the transformer model is competitive to standard LSTM based decoder in generating radiology report generation, in turn being significantly faster and parallelisable compared to their recurrent counterparts. Based on the extensive experiments by making changes on both encoder and decoder side, it proved the hypothesis that using state-of-the-art language models are beneficial in improving radiology report generation system.  
    
    \item \emph{Categorising concepts into common thoracic diseases as an auxiliary task}
    
    The radiology report generation is a complex and challenging problem. Given, we focus on generating reports from chest X-rays, it is important to capture domain specific image features for decoding. Physicians generally recommend chest X-ray studies to find common thoracic abnormalities such as \emph{pleural effusion}, \emph{emphysema}, and \emph{cardiomegaly}. In \citep{singh:2024:computeraideddiagnosisthoracicdiseases}, authors applied a multi-head self-attention augmented convolutional network for detecting fourteen diseases, which are the most common thoracic diseases as per the Fleischner society, on four of the large-scale datasets. Other studies such as ~\citep{Yuan:2019:Automatic_radiology_report} proposed a multi-task scheme having multi-label image classification as an auxiliary for radiology report generation. Experimental results have shown that this strategy provides better results because the encoder is enforced to learn radiology-related features. Therefore, detecting common thoracic diseases in chest X-rays benefits the radiology report generation. 
    
    \item \emph{Providing the CNN with additional information such as including both frontal and lateral projections from a chest X ray}
    
    Most of the radiology report generation systems considered only the posteroanterior (PA) or frontal view as large-scale publicly available chest X-rays datasets usually have that view. There are specific diagnosis cases where other views such as lateral (L) can complement the radiology study~\citep{Hashir:2020:quantifying_the_value_of_lateral_views}. There can be the case that lateral view has information which is not visible in the frontal view. Few studies, \citep{Hashir:2020:quantifying_the_value_of_lateral_views,Yuan:2019:Automatic_radiology_report} have investigated the importance of using multi-views as an input to the CNN model for image classification and radiology report generation. These studies indicated that adding multi-views such as frontal and lateral views can benefit the classification and report generation task. 
    
    \item \emph{The use of natural language generation and classification metrics}
    
    Automatic radiology report generation systems are evaluated based on machine translation, image captioning, and summarisation systems metrics, including BLEU~\citep{papineni-etal-2002-bleu}, CIDEr~\citep{Vedantam:2015:CIDEr}, METEOR~\citep{banerjee-lavie-2005-meteor}, and ROUGE~\citep{lin-2004-rouge}. These metrics calculate scores based on the overlap of n-grams in the reference text and the generated text. The generated radiology report should be well structured, coherent, clinically accurate, factually complete and consistent. In~\citep{singh:2024:clinicalcontextawareradiologyreport}, authors highlighted issues of using Natural Language Generation (NLG) metrics for evaluating radiology report generation systems. To re-iterate, NLG metrics are good at measuring overlap of words and fluency of text, however, they lack capabilities to evaluate the clinical accuracy, factual completeness, and consistency from diagnostic viewpoint. Research has been directed to novel metrics that can measure the diagnostic value in generated reports~\citep{Babar:2021:evaluating_diagnostic_content,Miura:2021:improving,Liu:2019:Clinically_accurate_CXR_generation}. Recently, \citep{Boag:2021:a_pilot_study} did a pilot study to investigate the clinical accuracy of generated reports as measured by machine translation and summarisation metrics. The authors highlighted that the NLG metrics are a proxy of model performance, and don't capture everything a radiologist would have when practising in real clinical settings. Recently, \cite{Yy:2023:evaluating_progress} investigated the correlation between automated radiology report generation metrics and radiologists. Authors highlighted limitations of existing metrics in capturing subtle medical differences and proposed \emph{RadGraph F1} and \emph{RadCliQ} so that there is a high match between radiologists and automated metrics. Based on investigations in~\citep{singh:2024:clinicalcontextawareradiologyreport}, as well as getting insights from existing studies in literature, we find the need to evaluate radiology report generation systems using both NLG metrics and diagnostic relevant metrics, as they complement each other.

    \item \emph{An interpretable and explainable radiology report generation}
    
    In high stake applications such as healthcare and defence, it is important that we should have have good understanding of how decisions are made from computer-aided systems. Given radiology report generation is an assistive system that can draft radiology reports augmenting radiologists, it is important to have interpretable and explainable radiology report generation systems so that radiologists and patients can trust these systems. Various techniques have been applied to add interpretability and explainability, such as Class Activation Maps (CAM)~\citep{Zhou:2016:CAM}, Grad-CAM~\citep{Selvaraju:2017:Grad-CAM}, Local Interpretable Model-Agnostic Explanations (LIME)~\citep{Ribeiro:2016:LIME}, Layer-Wise Relevance Propagation (LRP)~\citep{Bach:2015:LRP} and Concept activation vectors (CAV)~\citep{pmlr-v80-kim18d}.

\end{enumerate}

\section{Designing a robust radiology report generation system}
Based on the discussion of the above points, which are crucial and act as building blocks of a radiology report generation system, we finally outline a robust architecture to generate accurate and robust radiology reports from medical images.

\begin{figure}
    \centering
    \includegraphics[scale=0.5]{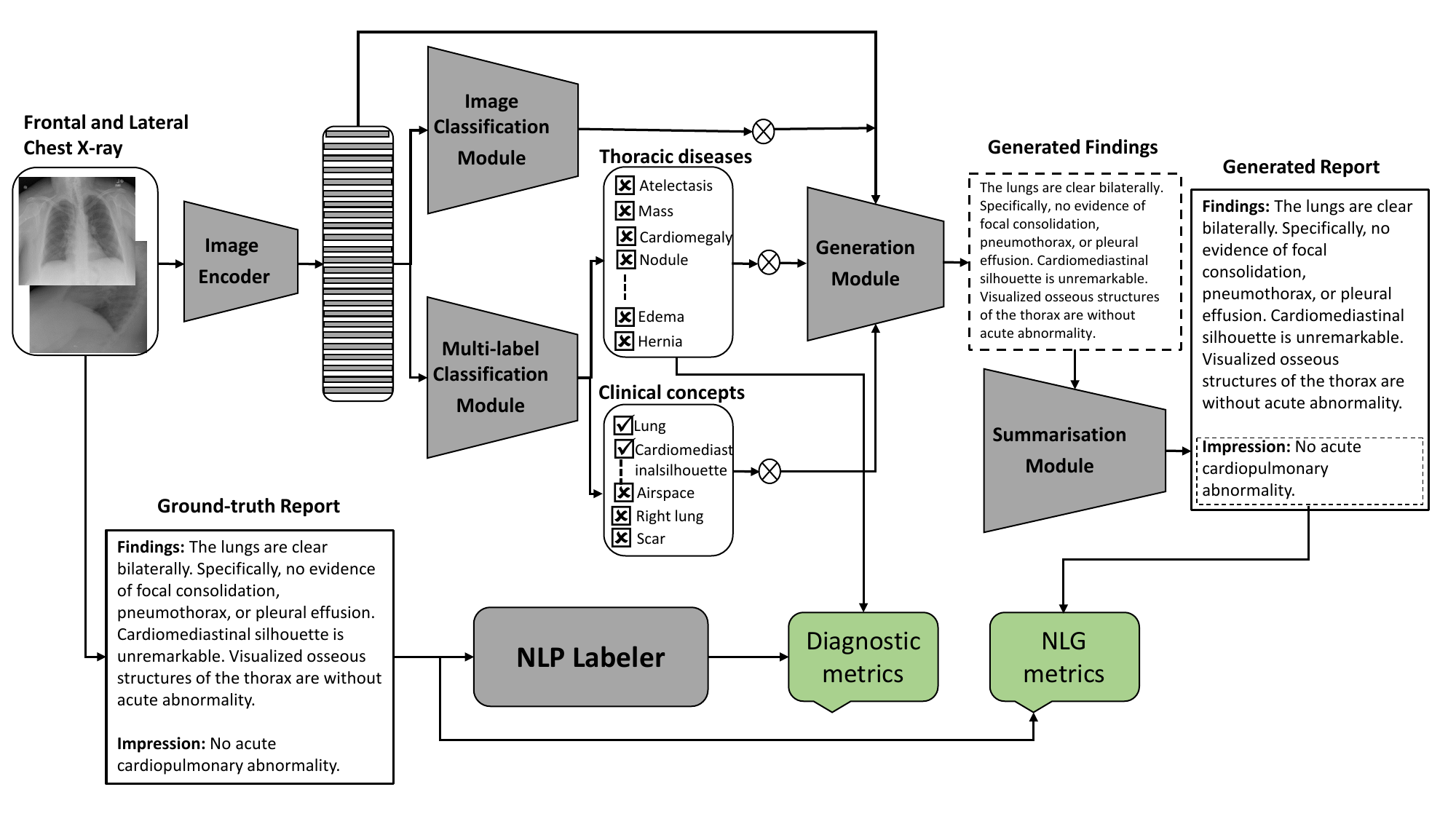}
    \caption{Block diagram of radiology report generation integrating multiple modules.}
    \label{fig:robust_radreport_generation_system}
\end{figure}

Figure~\ref{fig:robust_radreport_generation_system} shows the block diagram of an integrated system for robust radiology report generation. First, multi-view chest X-rays, having both the frontal and lateral views, are given as an input to the CNN to get image embeddings. The features of both the views are concatenated to form a single image representation. This image representation is used by the \emph{image classification module} and the \emph{multi-label classification module}. The \emph{image classification module} classifies an image as a \emph{normal} or \emph{abnormal}, which is important to separately generating findings for normal or abnormal images, as evident from study~\citep{Singh:2021:show_tell_summarise} that shows separating normal and abnormal images helps to improve model performance. The extracted image representation is further used by the \emph{multi-label image classification module} which performs two main tasks. First, it identifies the presence or absence of common thoracic diseases. Here, we focus on fourteen thoracic diseases, following the Fleischner Society's guidelines. The same has been followed in most of the large-scale publicly available datasets, namely, ChestX-ray14~\citep{Wang:2017:ChestX-ray14}, CheXpert~\citep{Irvin:2019:CheXpert}, and MIMIC-CXR~\citep{Johnson:2019:MIMIC-CXR}. Second, the multi-label classifier also identifies the presence of \emph{clinical concepts} or \emph{tags} in the input chest X-rays. Both the \emph{thoracic observations} and \emph{clinical concepts} are important to generate diagnostically accurate, complete, and consistent radiology reports. Once, we know the input chest X-rays as \emph{normal} or \emph{abnormal}, common \emph{thoracic observations}, and \emph{clinical concepts} present in chest X-rays, the \emph{findings generation module}, which is a \emph{decoder}, generates the \emph{findings section} for the given chest X-rays. The image classification module flags are used by the decoder to decide whether to generate findings for a \emph{normal image} or an \emph{abnormal image}. The generated findings are given as an input to the \emph{summarisation module} which summarises the findings section to give an \emph{impression section} as the output. The final generated report is the \emph{concatenation} of the findings and the impression sections, obtained as an output from the generation and summarisation modules respectively. In order to evaluate the performance of the integrated system for radiology report generation, we use both \emph{diagnostic metrics} and \emph{natural language generation metrics}, which provides robust measures of clinically accurate and coherent radiology reports. Specifically, we first applied \emph{CheXpert}, NLP labeler to get thoracic observations. The ground-truth labels of thoracic diseases (observations) are compared with the predicted labels of the \emph{multi-label classifier}, which provides a measure of how well the model is able to correctly identify the presence of common thoracic diseases. Further, the NLG metrics are also calculated by comparing the generated report with the ground-truth report. The first set of metrics, called \emph{diagnostic metrics}, are important to measure clinical accuracy and consistency among the generated report. Second, \emph{NLG metrics}, can measure whether the generated report is grammatically correct, fluent, and coherent. 

The overall system, built by integrating different modules proposed in this paper as well as using best practices from related studies, has the potential to generate clinically accurate and coherent radiology reports from chest X-rays. 

\section{Conclusion}

With rise of interest in apply multimodal learning to healthcare applications, the task of automatic radiology report generation aims at generating radiology reports from medical images by having comprehensive understanding of medical attributes, their relationship, and generating coherent, fluent, and diagnostically accurate reports. In this paper, we proposed a robust radiology report generation system by integrating best principles and practices based on the relevant literature. We discussed the impact of various methods and their benefits in an integrated system. The proposed architecture could generate clinically accurate radiology reports, in turn expediting diagnostic workflow, augmenting radiologists in decision making, and reducing diagnostic errors.

\section*{Declaration of competing interest}
The author(s) declare no conflicts of interest.




\vskip 0.2in
\bibliography{sample}

\end{document}